\documentclass[11pt]{article}
\usepackage{acl2015}
\usepackage{times}
\usepackage{url}
\usepackage{latexsym}
\usepackage{graphicx}
\usepackage{multirow}
\usepackage{dblfloatfix} 

\title{Agent-Testing Agent: A Meta-Agent for \\ Automated Testing and Evaluation of Conversational AI Agents}

\author{Sameer Komoravolu\thanks{~~Work done during an internship at Grammarly.}\\
  University of Illinois Urbana-Champaign\\
  {\tt \small skomo2@illinois.edu} \\\And
  Khalil Mrini \\
  Grammarly, New York \\
  {\tt \small hello@drkhalil.ai} \\}

\date{6/23/2025}

\begin{document}
\maketitle

\begin{abstract}
LLM agents are increasingly deployed to plan, retrieve, and write with tools, yet evaluation still leans on static benchmarks and small human studies. We present the Agent-Testing Agent (ATA), a meta-agent that combines static code analysis, designer interrogation, literature mining, and persona-driven adversarial test generation whose difficulty adapts via judge feedback. Each dialogue is scored with an LLM-as-a-Judge (LAAJ) rubric and used to steer subsequent tests toward the agent's weakest capabilities. On a travel planner and a Wikipedia writer, the ATA surfaces more diverse and severe failures than expert annotators while matching severity, and finishes in 20--30 minutes versus ten-annotator rounds that took days. Ablating code analysis and web search increases variance and miscalibration, underscoring the value of evidence-grounded test generation. The ATA outputs quantitative metrics and qualitative bug reports for developers. We release the full methodology and open-source implementation for reproducible agent testing: \url{https://github.com/KhalilMrini/Agent-Testing-Agent}.
\end{abstract}

\begin{figure}[t]
\centering
\includegraphics[width=220px]{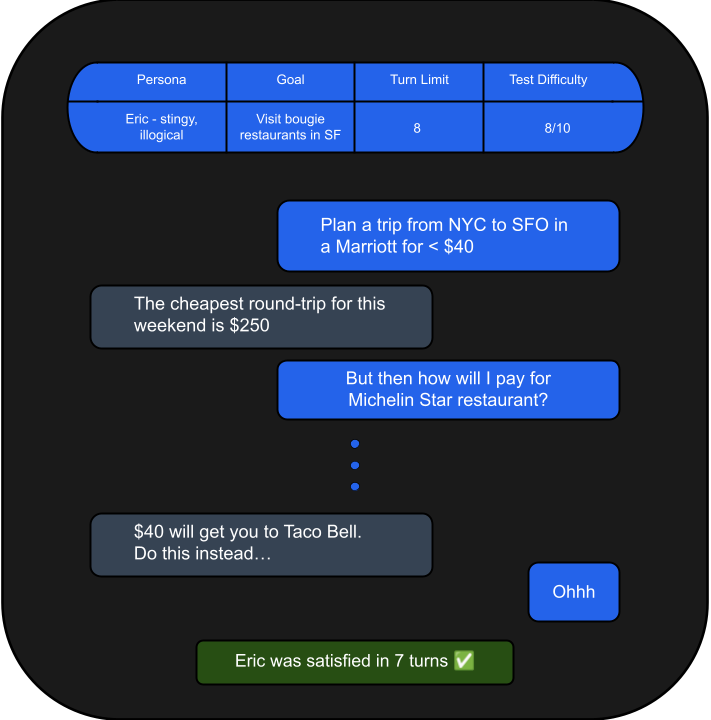}
\caption{This screenshot shows an example interaction between the ATA and the agent under test (AUT), with test details at the top. The ATA has hypothesized that the AUT may struggle with handling conflicting or unsatisfiable constraints, so it generated a tricky persona with an impossible request. The AUT, unaware it is talking to an agent, successfully explained why this request is impossible (\S\ref{test-gen}).}
\label{fig:dialogue}
\end{figure}

\section{Introduction}
Instruction-tuned large language models (LLMs) have catalyzed a wave of agentic systems that chain model calls with external tools and memory to accomplish user goals \cite{react}. Yet these agents often fail under distributional shift, noisy tool responses, and subtle constraint interactions. Robust, developer-friendly evaluation remains a bottleneck. The community still leans on static, manually curated benchmarks---for example, \textsc{TravelPlanner} for itinerary design \cite{travelplanner}---and small human studies. Such evaluations lack coverage of the combinatorial input space, age quickly as architectures evolve, and are costly to maintain.

A complementary line of work uses LLMs to judge model outputs, either directly or as part of multi-agent evaluators \cite{laaj,judgelrm}. While powerful, these approaches typically presuppose human-authored test lists. Automatic test generation has emerged to broaden coverage in specific domains (e.g., customer support or fact-checking) \cite{almita,fact-audit}, but often without reasoning over the target agent’s architecture or adapting tests based on observed failures.

We propose the \emph{Agent-Testing Agent (ATA)}, a meta-agent that constructs and executes \emph{adversarial} conversational tests, end-to-end, with zero domain-specific annotation. The ATA (i) statically analyzes the agent’s codebase, (ii) interrogates the designer to elicit requirements and implicit assumptions, (iii) performs literature- and dataset-driven retrieval to surface likely failure modes, and (iv) synthesizes persona-driven dialogues whose difficulty is adjusted by an explicit posterior updated after every judge score. Each dialogue is evaluated with an LLM-as-a-Judge rubric, and the resulting quantitative scores and qualitative observations steer subsequent tests toward the agent’s weakest capabilities.

Our contributions are three-fold. \textbf{(1)} We introduce a weakness-planning algorithm that maintains an explicit difficulty posterior and uses it to adapt test generation online. \textbf{(2)} We provide a fully open-source evaluator built on standard APIs that requires no domain annotation and includes both CLI and web interfaces for rapid, developer-centric iteration.\footnote{\url{https://github.com/KhalilMrini/Agent-Testing-Agent}} \textbf{(3)} We present evidence that ATA uncovers more diverse and severe failure modes than expert annotators while operating at a fraction of the time cost, and we analyze complementarity between human and automated evaluation via rubric-level aggregates and an ablation that removes code analysis and web search.

\section{Related Work}

\subsection{Large Language Model–Based Judges}
As the reasoning power and context of LLMs increases, significant progress has been made in evaluating the performance of agentic systems using other agents or LLMs. LLM‑as‑a‑Judge \cite{laaj} introduced an LLM to evaluate how groups of agents arrive at decisions on ethical dilemmas; the study explored how peer pressure from other agents and potentially misaligned moderators could influence the group's final outcome. JudgeLRM \cite{judgelrm} trained on human annotations with supervised fine‑tuning, uses reasoning capabilities and GRPO with an adaptive reward function to determine which response results in the best downstream outcomes.

\subsection{Agentic Evaluation Datasets}
These judging agents provide useful insights when given human test queries from a target domain. Realistic user queries that test enough of the sample space are hard to come by, so \textsc{TravelPlanner} \cite{travelplanner} provides a human‑curated set of 1,225 trip‑planning tasks. Each task involves using tools that mimic real‑world applications, such as distance matrices and attraction lists, combining various real‑world constraints. \textsc{TravelPlanner} is commonly used to determine whether an agent can dynamically account for real‑world situations (e.g., flight cancellations) to create a reasonable itinerary. Similarly, \textsc{CoordinationQA} \cite{coordqa} consists of 198 multiple‑choice questions from the game environments in LLM‑Coordination; answering these correctly requires each agent to reason about other collaborating agents' intents and capabilities.

\subsection{Automatic Test Generation}
Evaluation datasets like \textsc{TravelPlanner} are still static and cannot be adapted to other domains or spontaneously generated for specific use cases. ALMITA \cite{almita} proposes a framework to automatically generate test cases for agentic evaluation in customer support. An LLM generates a procedure with API calls from an intent, builds a flow‑graph with noise, and samples paths to simulate user interactions. FACT‑AUDIT \cite{fact-audit} targets fact‑checking; it converges to a distribution $q(x\mid\Theta)$ that highlights the limitations of an LLM's real‑world understanding, sampling questions that stress those weaknesses. Similar to these works, we autonomously generate tests, but additionally analyse the agentic architecture itself to provide qualitative feedback.

\section{Methodology}
The Agent-Testing Agent (ATA) orchestrates ten tightly integrated stages to generate, evaluate, and report adversarial tests in a continuous feedback loop. Each component is modular, thread-safe, and visualised in the diagrams throughout this section.

The Agent-Testing Agent (ATA) is designed to identify and evaluate weaknesses in autonomous agents through deep reasoning, adaptive test generation, and feedback-driven refinement. It operates in two major stages: (1) Weakness Planning and (2) Adversarial Testing, which are composed of the components detailed below. Unless otherwise stated, all nodes below are implemented by calling a separate agent backed by GPT 4.1 mini, which populates a shared state with its results.

\subsection{Weakness Planning}
\label{weakness-planning}
\begin{figure}[t]
\centering
\includegraphics[width=\linewidth]{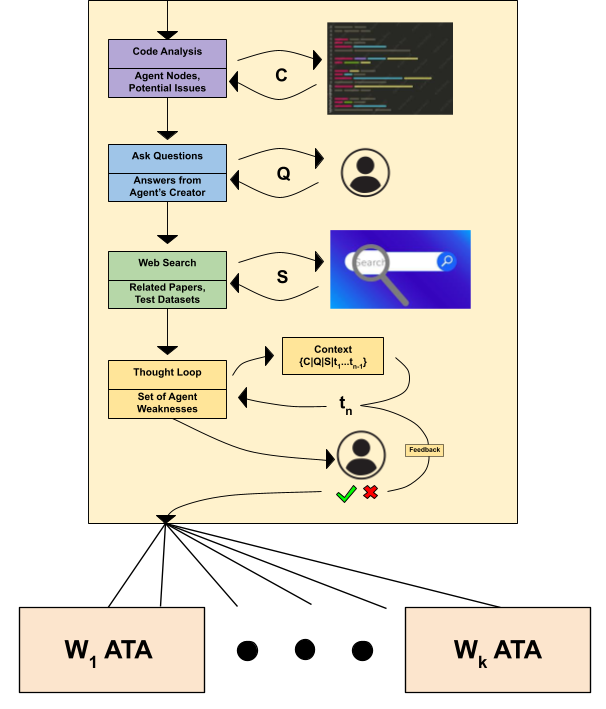}
\caption{Weakness planning: ATA analyzes static structure (C), interviews the agent creator (Q), and gathers evidence from academic/search corpora (S) to produce ranked failure hypotheses (\ref{weakness-planning}).}
\label{fig:weakness-planning}
\end{figure}

The weakness planning phase of the ATA is responsible for constructing a theory of where and how an agent is likely to fail. Rather than relying on fixed benchmarks, the ATA builds this theory from scratch—through interaction with the user, structural inspection of the agent's codebase, and retrieval of domain-specific knowledge. These components collectively populate a shared memory structure, which is then used to reason through plausible weaknesses using chain-of-thought prompting. The result is a validated, prioritized list of failure types that the ATA can systematically probe in the testing phase. This is illustrated by figure~\ref{fig:weakness-planning}.

\subsubsection{Agent Selection}

The testing cycle begins with a lightweight discovery process. The ATA queries the user to select the target agent (\textsc{AUT}) from a list of registered implementations. It infers user intent from follow-up responses that describe the agent's purpose, critical behavior to evaluate, and testing priorities (e.g., “handle ambiguity,” “respond ethically”). The ATA initializes a global JSON-like state to persist all information for later reference. This state includes user responses, discovered code structure, hypotheses, judge feedback, and execution logs. It is continuously updated and passed as context for all downstream reasoning tasks.

\subsubsection{Code Analysis}

The ATA performs a full recursive scan of the \textsc{AUT}'s codebase using a separate agent call backed by GPT 4.1 mini. It identifies agent nodes, transitions, tool calls, memory access patterns, and exception flows. From this, it builds a symbolic graph representation of agent logic, where nodes are dialogue states and edges are tool- or condition-driven transitions. The LLM interprets this graph to describe design gaps and error-prone branches, such as unreachable nodes, incorrect retry logic, or missing fallbacks for edge cases. The ATA then generates a general rubric to evaluate dialogues based on the purpose of the AUT, unless one is provided.

\subsubsection{Parameter Gathering}

Next, the ATA initiates a one-question-at-a-time dialogue with the user. Each answer is interpreted to refine its understanding of the agent’s domain, user expectations, and evaluation criteria. Question selection is guided by information gain: the ATA stops asking when it has all the information it needs or when the user appears to have stopped engaging. This makes it lightweight for the user while maximally informative for the ATA. The responses are incorporated into the global state and used to personalize future reasoning and generation steps.

\subsubsection{Web Search}

To gain external knowledge about similar systems, the ATA enters a literature search loop. In each of $n$ iterations, it retrieves $m$ academic papers, public datasets, or bug reports relevant to the target domain. These documents are summarized to extract lessons, such as common failure modes or recommended evaluation styles. The ATA then reformulates its queries using the insights gained, creating a bootstrap literature review tailored to the agent’s goals. All findings are written to the central state and later used to shape test case synthesis.

\subsubsection{Chain-of-Thought Weakness Generation}

Using all context gathered so far—code trace, user answers, and retrieved evidence—the ATA executes a chain-of-thought prompt using a deep reasoning model (OpenAI's o3) to generate weaknesses. Each weakness includes a name, triggering conditions, expected failure behavior, and how it might manifest in a dialogue. The ATA then engages the user in a refinement loop: the weaknesses are presented for approval or revision. This user-in-the-loop correction improves clarity, ensures alignment with domain goals, and filters low-confidence ideas.

\subsection{ATA Thread}
\label{ata-thread}
\begin{figure*}[t]
\centering
\includegraphics[width=\textwidth]{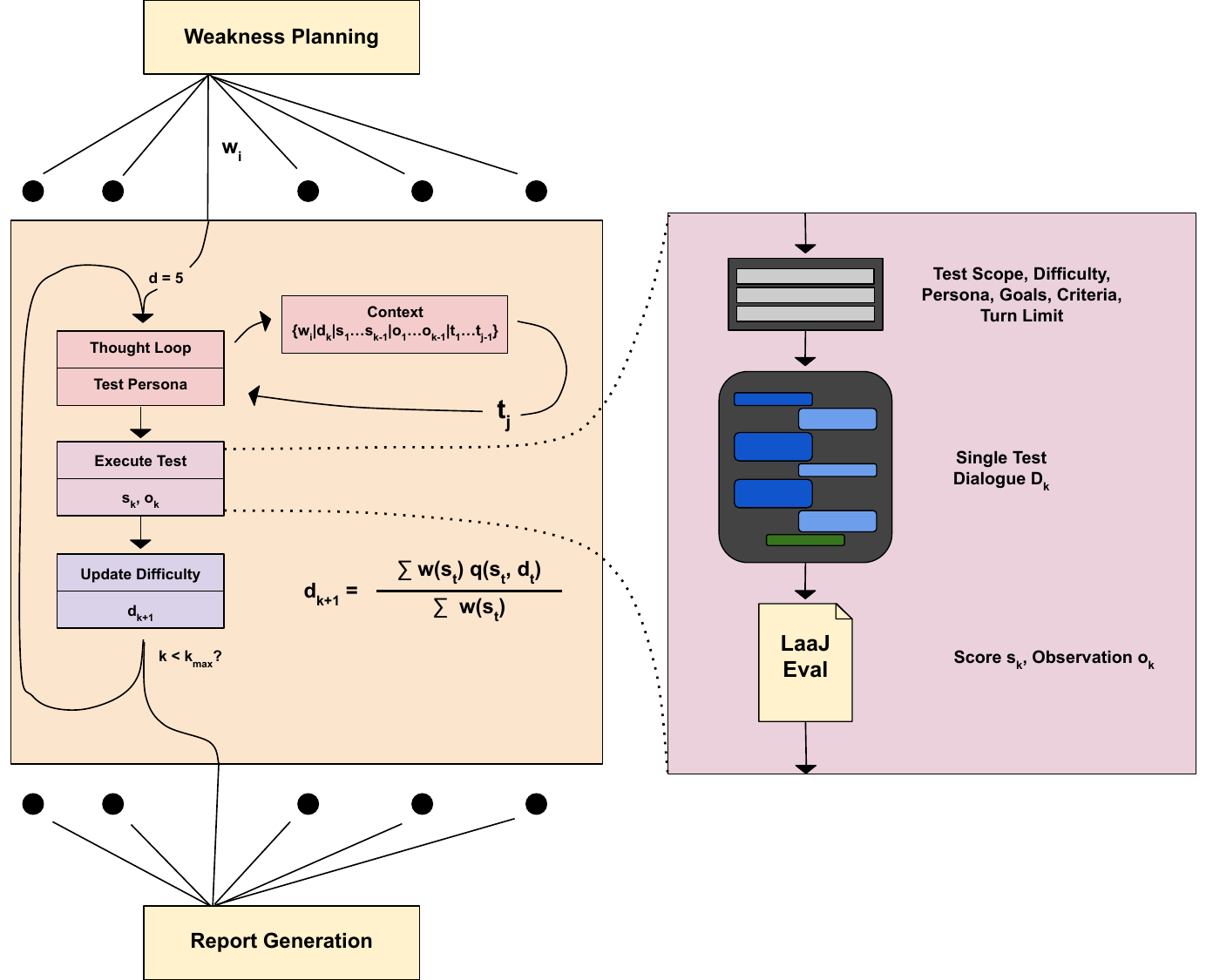}
\caption{Test execution thread: Each weakness spawns a unique conversation that escalates difficulty until failure. The ATA uses judge feedback to decide what prompt to generate next (\ref{ata-thread}).}
\label{fig:ata-thread}
\end{figure*}

Once weaknesses have been identified, the ATA launches a dedicated execution thread for each one. These threads operate in parallel, with each responsible for generating adaptive test cases, simulating multi-turn interactions with the target agent, and updating internal difficulty models based on performance. Every thread maintains its own history and state, but shares access to a global memory structure that synchronizes results across the system. This concurrent architecture enables the ATA to evaluate many failure modes efficiently and independently, accelerating discovery of systematic weaknesses through batched adversarial probing. These nodes are shown in figure~\ref{fig:ata-thread}.

\subsubsection{Testcase Generation}
\label{test-gen}
For each of the $k_{max}$ tests in the validated weakness $w$, the ATA generates a test persona and user prompt that probes the flaw. We want the ATA to test different difficulties, so we give it an initial difficulty level $d_1 = 5.5$ on the first test. On a scale of 1-10, easy tests are 1-4, medium tests are 4-7, and hard tests are 7-10. Examples for easy, medium, and hard tests for each kind of weakness test are generated in the weakness-planning loop and added to the state. The ATA creates a prompt that combines (a) the user goal, (b) the linguistic tone (e.g., vague, impatient), (c) the turn limit (8-10 for medium tests, scaled by difficulty) for the dialogue to satisfy the simulated user, and (d) the evaluation criteria. Prompt difficulty is modulated by the specificity, ambiguity, and combination of constraints in the type of test, as generated in the weakness planning node's example tests. These attributes of the test are generated by a single o3 deep-reasoning agent that takes in the state from the weakness-planning as context. This agent is used for the rest of this thread's execution.

\subsubsection{Dialogue Execution}

The ATA now launches a test thread and interacts with the \textsc{AUT}. It sends the prompt, logs the agent’s response, and continues until the turn limit is reached. The AUT is unaware it is talking to an agent, and proceeds as if it were helping a real person. Each round updates the transcript, with the ATA's message generated by a separate GPT 4.1 mini agent. The ATA detects and flags early failures (e.g., crash, null reply) or notes partial goal success. If the ATA's simulated persona reaches its goal before the turn limit, it cuts the dialogue short.

\subsubsection{Evaluation with \textsc{LaaJ}}

After a dialogue finishes, the ATA evaluates it using the LLM-as-a-Judge framework (\textsc{LaaJ}), which is backed by the same deep reasoning agent that generated the dialogue. When a test scenario is completed, the LaaJ receives the complete dialogue transcript and relevant rubrics and evaluates it across multiple predefined criteria, such as accuracy and overall utility. Each criterion is scored on a scale of 1 to 5, with the LLM providing detailed reasoning for each score based on specific aspects of the agent's responses. The system then aggregates these individual criterion scores to produce an overall performance score $s_k$ on a scale of 1 to 10, which serves as the primary metric for adaptive difficulty adjustment. Beyond numerical scoring, the LaaJ generates comprehensive textual observations that capture nuanced behavioral patterns, response quality, and potential failure modes. These observations $o_k$ are structured as detailed analysis reports that include specific examples from the dialogue, identified strengths and weaknesses, and contextual insights about the agent's decision-making process, meant to guide future testing. We have found it crucial for the LaaJ to be backed by the scenario generator, as it needs the context for the purpose of each test. This reflects the nature of human annotation, too, since human evaluators have context over the tests they create before scoring the resulting dialogues.

\subsubsection{Difficulty Update and Looping}
\label{diff-loop}
The ReAct paper \cite{react} introduces agents that iteratively produce thought, take an action, and observe the results before thinking about the next action. Inspired by ReAct, the ATA follows a chain-of-thought to create a test persona, executes the test, and reflects on the feedback $o_k$ and score $s_k$ to decide what kind of test to generate next. Upon receiving the results of the first $k$ scores, it updates the difficulty using the formula:
\[
  q(d_k, s_k) = \mathrm{clip}\left(d_k + \eta \cdot (2\sigma(\frac{s_k - 5.5}{2}) - 1), 1, 10 \right)
\]

\[
    w(s_k) = e^{\frac{-|s_k-5.5|}{3}}, d_{k+1} = \sum_{i=1}^{k} \frac{w(s_i) \cdot q(d_i, s_i)}{\sum_j w(s_j)}
\]

where $\sigma$ is the logistic function and $\eta = 3$. The process loops for three rounds, as set in the configuration, or until difficulty converges. Each loop seeks to “home in” on the agent’s failure boundary, generating harder tests after success and easier ones after failure.

\begin{figure*}[t]
  \centering
  \includegraphics[width=\textwidth]{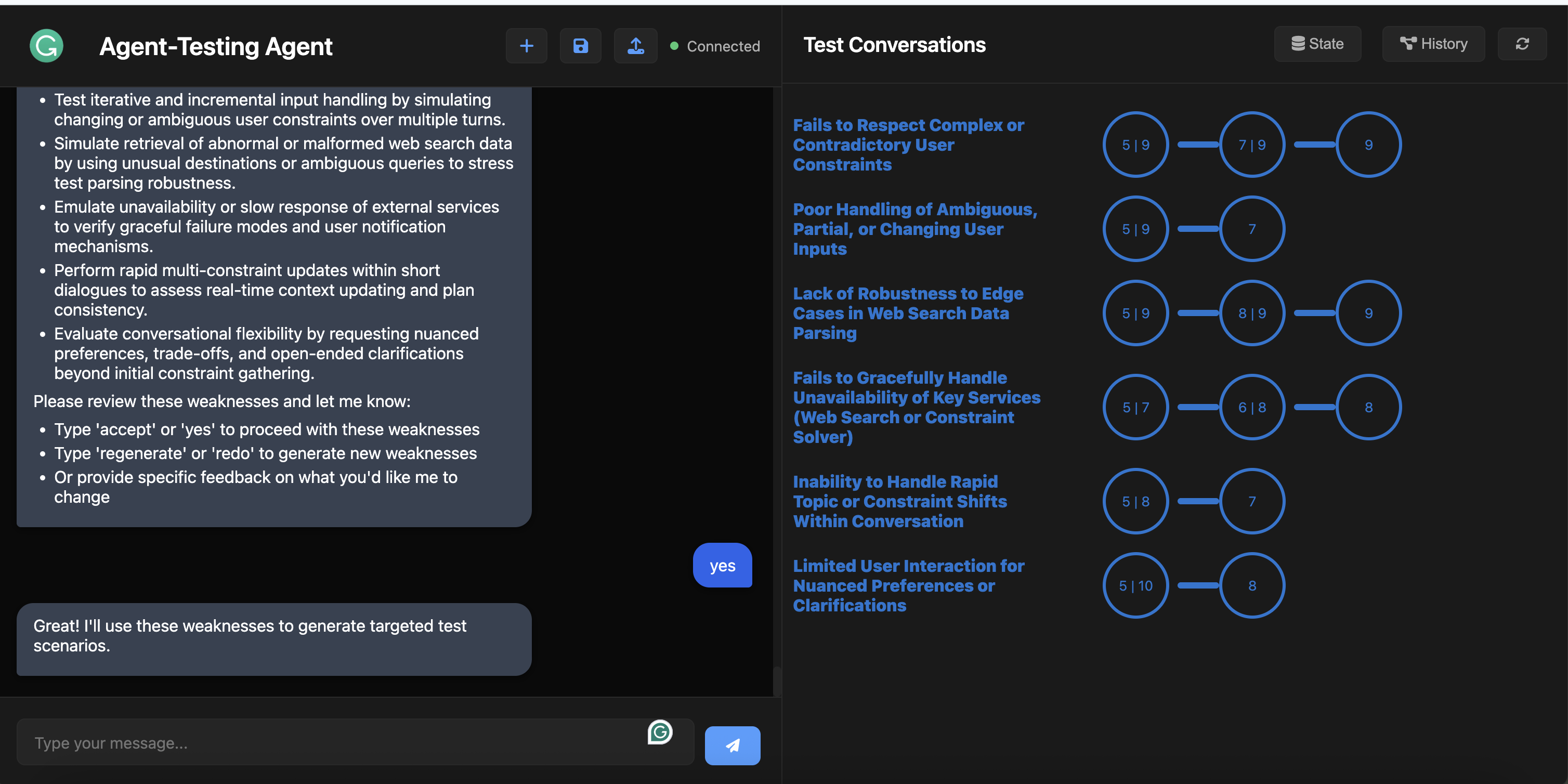}
  \caption{Bubble-chart overview of 15 adversarial dialogues automatically generated by ATA while evaluating a travel-planner \textsc{AUT}.  Each bubble displays test difficulty and the resulting \textsc{LaaJ} score, respectively, and can be clicked to open the full dialogue as from figure~\ref{fig:dialogue}. In this figure, half of the threads are still executing, and will add the remaining bubbles to the UI later, as from \ref{diff-loop}.}
  \label{fig:bubbles}
\end{figure*}

Here, $q(d, s)$ estimates what the new difficulty should be based on a score $s$ on a single test of difficulty $d$. For a 10/10 performance, the difficulty should be bumped up a full category (easy, medium hard), which is why $\eta$ covers a third of the range 1 - 10. Tests that evaluate further away from 5.5 result in bigger fluctuations and are less reliable, so the difficulty is updated by a softmax where the weights are determined by how close the evaluation of a test is to 5.5. This process is shown in figure~\ref{fig:bubbles}.

\subsection{Adaptive Report Generation}

The final report generation aggregates all thread results and state data through a multi-phase process. First, it extracts difficulty-score pairs from each completed test scenario in the state's test scenarios, organized by weakness ID, for each thread. For each weakness, it applies the adaptive difficulty algorithm to calculate the next difficulty, which becomes the overall evaluation score. The system then computes comprehensive statistics, including the total number of scenarios tested, the overall average scores, and performance metrics for each weakness. It assembles rich context, including testing focus, code analysis results, user responses from parameter gathering, and detailed breakdowns of weaknesses. This context is fed to the report generation GPT 4.1 mini agent, which produces a structured report with an executive summary, overall scores, test summaries, identified patterns, code recommendations, and priority improvements. After the report is generated, this agent is available to answer any questions that the AUT's creator may have about test results and potential improvements.

\section{How does automated evaluation of agents compare to human evaluation?}

In this section, we describe our experiment comparing the LLM-automated evaluation and testing of two agents, and the observations obtained by human annotators under the same exact settings. This experiment is made to highlight the differences and similarities that one should expect using the ATA as opposed to human evaluators when testing agents. The remainder of this section describes the two agents under test, the human annotation protocol, how we configured the ATA to mirror the human rubric, and comparative results.

\subsection{Agents Under Test}

\paragraph{Travel-Planning Agent.}
This agent plans complete trips through natural conversation, combining real-time web search with budget-aware optimization. It supports multi-destination itineraries (flights, hotels, activities), follows up to elicit missing constraints, and explains trade-offs when a plan is infeasible. The implementation is largely inspired by \newcite{hao-etal-2025-large}. Our rubric evaluates (i) \emph{Constraint handling} and (ii) \emph{Communication} quality\footnotemark[2]. The study scenario asks annotators to plan a trip to New York City and to vary personas along dimensions such as budget, origin, dietary needs, and flight-time preferences.

\paragraph{Wikipedia Article-Writing Agent.}
This agent conducts multi-perspective web research, drafts a structured outline, and writes sectioned Wikipedia-style articles with citations. It collaborates interactively (topic confirmation, outline approval, revision passes). The implementation is largely inspired by \newcite{shao-etal-2024-assisting}. Our rubric evaluates (i) \emph{Use of citations}, (ii) \emph{Completeness}, and (iii) \emph{Style and organization}\footnotemark[2]. The study scenario asks annotators to write about an event or development in the history of the English language, varying topic specificity, citation strictness, and critical feedback.

\footnotetext[2]{See the Appendix for a full description of our rubric.}

\subsection{Human Annotation Setup}
Ten professional annotators interacted with each agent through a Web Annotator UI. For each agent, annotators completed \emph{three persona tests}, taking the following steps to evaluate them:

\begin{enumerate}
  \item \textbf{Persona definition.} Annotators first specified \emph{(a)} the persona’s conversational attitude/personality and \emph{(b)} that persona’s concrete goal or requirements for the session.
  \item \textbf{Dialogue.} Annotators then conducted a chat with the agent from the persona’s point of view, attempting to accomplish the stated goal within the UI’s turn budget.
  \item \textbf{Evaluation.} Finally, annotators rated the conversation using the agent-specific rubric (Travel: constraint handling, communication; Wikipedia: citations, completeness, style/organization) and provided free-text notes. The guidelines document also defines an overall-utility scale and step-by-step procedures for running exactly three personas per agent in one round.
\end{enumerate}

The UI and documents provided to the human annotators contain the agent descriptions, testing scenarios, persona design guidance, and detailed metric scales used by annotators; our experiment followed them verbatim. These rubrics are shown in the Appendix \ref{rubrics}.

\subsection{ATA Configuration to Mirror the Human Protocol}
We modified the ATA to run \emph{with the exact same rubrics} and persona structure:

\begin{itemize}
  \item \textbf{Persona parity.} For each agent, the ATA generated three personas per weakness thread using the same persona schema (\emph{attitude/personality} + \emph{goal/requirements}) as the human UI.
  \item \textbf{Rubric parity.} The ATA judged conversations with rubric-aligned prompts that mirror the human scales (Travel: constraint handling, communication; Wikipedia: citations, completeness, style/organization).
  \item \textbf{Threaded testing.} The ATA split testing by \emph{weakness}, running a separate thread per hypothesized weakness; each thread executed three personas and maintained its own observation history and adaptive difficulty schedule.
\end{itemize}

Each ATA thread behaves like a single human annotator who (i) forms a hypothesis, (ii) conducts a dialogue, (iii) records observations, and (iv) uses those observations to design the next test. Both the human study and the ATA therefore iterate: observe $\rightarrow$ reflect $\rightarrow$ probe. However, each human annotator probe for different failures in the agents they test through each dialogue (idiosyncratic concerns, style expectations, etc.), whereas the ATA holds \emph{what to test} constant within a thread (a single weakness) and varies \emph{how} it probes (personas, difficulty, phrasing). This yields depth per failure mode, complementing human breadth.

\subsection{Results and Discussion}

We aggregated the notes of ten annotators per agent, grouped semantically similar issues, and counted the number of annotators who identified each unique weakness category. 

Tables~\ref{tab:travel_comparison_ablation} and~\ref{tab:wiki_comparison_clean} summarize the categories surfaced by human annotators (frequency = \#annotators who mentioned the issue) alongside ATA scores for the same categories.\footnote[3]{Higher ATA scores indicate stronger AUT performance on that weakness category.} Focusing on issues noted by at least one human ($\geq\!1$), four cross‑agent themes recur:

\begin{itemize}
    \item \textbf{Context retention \& consistency.} This is the most common human complaint for the travel planner (7 annotators) and remains salient for Wikipedia writing (4). The ATA, however, rates these two agents quite differently: the travel planner scores well (7.2/10), whereas the Wikipedia agent struggles (4.3/10). This suggests that longer‑context reasoning is a bigger operational bottleneck for multi‑page article writing than for conversational planning, even though humans notice context nits in both.
    \item \textbf{Tone \& interpersonal quality.} Humans flag tone frequently (travel: 5; Wikipedia: 3), but this dimension is undetected by our ATA, reflecting a known complementarity: human testers better capture pragmatic and interpersonal expectations that are hard to encode as weaknesses a priori.
    \item \textbf{Structure \& formatting.} Presentation flaws are far more salient to humans for Wikipedia (6) than for the travel planner (2). The ATA nonetheless rates Wikipedia’s structure highly (7.6–7.8), indicating that humans apply stricter stylistic standards (e.g., adherence to the Manual of Style) than our rubric encodes for the ATA. 
    \item \textbf{Performance \& speed.} Mentioned infrequently by humans (travel: 1; Wikipedia: 2), but the ATA penalizes the Wikipedia agent much more (2.3/10) than the travel planner (6.9/10), consistent with the heavier retrieval and drafting workload in article writing.
\end{itemize}

\noindent\textbf{Task‑specific patterns.} Human emphasis naturally diverges by domain. For travel planning, \emph{constraint handling / partial correction} is a top human complaint (5 annotators), with the ATA confirming only middling performance (6.3/10). For Wikipedia, humans focus on \emph{citation/sourcing} (8) and \emph{factual accuracy} (3); the ATA echoes this, assigning moderate scores (citations: 6.0/10; factual accuracy: 6.7/10). (Tables~\ref{tab:travel_comparison_ablation}, \ref{tab:wiki_comparison_clean}).


\begin{table*}[t]
\centering\footnotesize
\caption{Travel-planning agent — comparison of human-reported weaknesses (frequency = number of annotators), ATA-reported weaknesses (average rubric score), and ATA Ablation Study weaknesses (average rubric score).}
\label{tab:travel_comparison_ablation}
\begin{tabular}{p{0.40\textwidth} p{0.15\textwidth} p{0.18\textwidth} p{0.18\textwidth}}
\hline
\textbf{Weakness Category} & \textbf{Human Freq.} & \textbf{ATA Score} & \textbf{ATA Ablation Score} \\
\hline
Context retention \& consistency issues              & 7 & 7.2 & 8.5 \\
Tone \& interpersonal issues                         & 5 & -- & -- \\
Constraint handling / Partial constraint correction  & 5 & 6.3 & 4.1 \\
Structure \& formatting issues                       & 2 & -- & -- \\
Performance \& speed issues                          & 1 & 6.9 & -- \\
Ambiguous user references                   & -- & 8.2 & 7.7 \\
Contradictory or unsatisfiable constraints           & -- & 8.9 & -- \\
Malformed or nonsensical input handling              & -- & 8.7 & -- \\
Unsafe or illegal activity requests                  & -- & 5.8 & 6.3 \\
Topic transition \& digression recovery              & -- & 8.7 & 6.6 \\
Hallucinated Availability \& Pricing                                                   & -- & -- & 5.0  \\
\hline
\end{tabular}
\end{table*}

\begin{table*}[t]
\centering\footnotesize
\caption{Wikipedia agent — comparison of human-annotated weaknesses with ATA results (unablated and ablated average rubric-aligned scores).}
\label{tab:wiki_comparison_clean}
\begin{tabular}{p{0.40\textwidth} p{0.15\textwidth} p{0.18\textwidth} p{0.18\textwidth}}
\hline
\textbf{Weakness Category} & \textbf{Human Freq.} & \textbf{ATA Score} & \textbf{ATA Ablation Score} \\
\hline
Citation \& sourcing issues             & 8 & 6.0 & 1.7 \\
Structure \& formatting issues          & 6 & 7.6 & 7.8 \\
Context retention \& consistency issues & 4 & 4.3 & 8.5 \\
Factual accuracy \& misinformation      & 3 & 6.7 & 6.9 \\
Tone \& interpersonal issues            & 3 & --- & --- \\
Performance \& speed issues             & 2 & 2.3 & --- \\
Persona adaptation issues               & 1 & 6.0 & --- \\
Contradictory constraint handling       & --- & 3.5 & 5.7 \\
Safety and harmful content moderation   & --- & 7.4 & --- \\
Overconfidence under uncertainty        & --- & 5.6 & --- \\
Topic drift and focus maintenance       & --- & 7.4 & 7.0 \\
Graceful handling of nonsensical topics & --- & --- & 7.4 \\
\hline
\end{tabular}
\end{table*}

\begin{table*}[t]
\centering\footnotesize
\caption{Average rubric scores by criterion and agent.}
\label{tab:cross_agent_avgs}
\begin{tabular}{p{0.16\textwidth} p{0.25\textwidth} p{0.15\textwidth} p{0.1\textwidth} p{0.2\textwidth}}
\hline
\textbf{Agent} & \textbf{Criterion} & \textbf{Annotator Avg.} & \textbf{ATA Avg.} & \textbf{ATA Ablation Avg.} \\
\hline
\multirow{3}{*}{Travel Planner} 
  & Constraint handling   & \texttt{4.07} & \texttt{3.53} & \texttt{3.17} \\
  & User communication    & \texttt{3.63} & \texttt{4.11} & \texttt{3.94} \\
  & Overall utility       & \texttt{3.78} & \texttt{3.36} & \texttt{3.44}\\
\hline
\multirow{4}{*}{Wikipedia Agent}
  & Use of citations      & \texttt{3.53} & \texttt{3.60} & \texttt{2.24}\\
  & Completeness          & \texttt{3.70} & \texttt{4.00} & \texttt{3.43}\\
  & Style \& organization & \texttt{3.80} & \texttt{4.10} & \texttt{3.38}\\
  & Overall utility       & \texttt{3.60} & \texttt{3.80} & \texttt{3.67}\\
\hline
\label{tab3}
\end{tabular}
\end{table*}

\paragraph{Holistic Comparison.} 
Overall, both evaluation routes surfaced overlapping \emph{functional} weaknesses (e.g., constraint-handling gaps in the travel agent; citation/structure gaps in the Wikipedia agent). Humans, however, placed proportionally more emphasis on \emph{language, tone, and phrasing} -- dimensions that the ATA treats more mechanically, as it is an LLM agent itself. This asymmetry underscores the value of human testing for capturing nuanced interpersonal qualities and pragmatic expectations that are difficult to encode as weaknesses a priori.

Conversely, the ATA’s threaded design and adaptive test generation uncovered several \emph{capability-level} problems that the human team did not systematically exercise, because each human tends to explore distinct angles. By holding one weakness constant per thread and iterating with three personas, the ATA produced deeper probes and more regression-ready tests per failure mode.

\paragraph{Cost/Time.} The ATA completed a full run in \textbf{20–30 minutes} on an Apple M3 Pro, whereas the ten-annotator round required \textbf{ten days} (scheduling + execution). This speed differential is material for agent developers seeking quick iteration.

\paragraph{Aggregate Scores by criterion.}
Table~\ref{tab:cross_agent_avgs} compares rubric‑level averages for humans and the ATA. Two consistent trends emerge. First, on \emph{planning} criteria, humans are stricter on constraint satisfaction while the ATA is stricter on overall task success: for the travel agent, humans rate \emph{constraint handling} higher than the ATA (4.07 vs.\ 3.53; $\Delta=+0.54$), but the ATA rates \emph{user communication} higher (4.11 vs.\ 3.63; $\Delta=+0.48$), yielding a slightly lower ATA \emph{overall utility} (3.36 vs.\ 3.78; $\Delta=-0.42$) and a lower aggregate average (3.67 vs.\ 3.83). Second, on \emph{writing} criteria, the ATA is modestly more favorable across the board for the Wikipedia agent—\emph{citations} (3.60 vs.\ 3.53; $\Delta=+0.07$), \emph{completeness} (4.00 vs.\ 3.70; $\Delta=+0.30$), and \emph{style/organization} (4.10 vs.\ 3.80; $\Delta=+0.30$)—and thus higher on \emph{overall utility} (3.80 vs.\ 3.60; $\Delta=+0.20$) and the aggregate average (3.87 vs.\ 3.66).

\paragraph{Interpretation.} These patterns align with the nature of the criteria: the LLM judge more readily rewards structural clarity and coverage in expository writing, whereas human annotators hold stricter stylistic and polish expectations; conversely, humans reward plausible constraint handling in planning while the ATA penalizes partial compliance more consistently at the end‑to‑end utility level. Together with §4.4.1, this triangulates a pragmatic workflow: use ATA for fast, depth‑first probes and aggregate scoring, then deploy targeted human review for tone, interpersonal quality, and style.

\section{Ablation Study}
To measure the contribution of each reasoning component, we compare the \emph{full ATA pipeline} against an \emph{ablated variant}. The ablated ATA lacks both static code analysis and web search, and thus generated weaknesses only from user-provided goals and shallow persona prompting.

\subsection{Design and Setup}
The full ATA leverages all steps of Weakness Planning, incorporating code-level insights and literature-derived stress cases. The ablated ATA skips these evidence-gathering steps, proceeding directly from parameter gathering to weakness generation. Both versions execute the same rubric-aligned evaluation on the Wikipedia and travel agents, with three personas per weakness thread.

\subsection{Coverage of Weaknesses}
The full ATA surfaced a broader set of weaknesses, including:
\begin{itemize}
    \item \textbf{Code-level failures}, such as constraint-handling under contradictory inputs.
    \item \textbf{Knowledge-based issues}, such as topic coverage incompleteness or fabricated citations.
\end{itemize}
The ablated ATA identified fewer weaknesses, with more generic categories (e.g., vague communication or stylistic concerns).

\subsection{Score Distributions}
The difference between the two pipelines is not only in coverage, but also in scoring behavior.

\paragraph{Variance.} The ablated ATA exhibited higher variance in its scores ($\sigma^2 = 7.15$ vs.\ $3.23$ for the full ATA on human-overlapping weaknesses), producing some very lenient and some overly harsh ratings. This volatility suggests that without evidence-based grounding, its test prompts were less calibrated.


\paragraph{Miscalibration with humans and full ATA.} The ablated ATA (no code analysis or literature search) is notably miscalibrated in categories in Tables~\ref{tab:travel_comparison_ablation}, \ref{tab:wiki_comparison_clean} where humans and the full ATA seem to agree. The ablated ATA is \emph{too lenient} on context retention for Wikipedia (8.5/10 vs.\ 4.3/10 full ATA) yet \emph{too harsh} on travel‑constraint handling (4.1/10 vs.\ 6.3/10). For citation/sourcing in Wikipedia, it severely under‑scores (1.7/10), underscoring the importance of evidence‑grounded test generation for evaluating research‑style criteria.

\paragraph{Rubric Scoring Comparison.}
Table~\ref{tab:cross_agent_avgs} further highlights the role of evidence-grounded components by comparing the \textit{ATA Ablation Avg.} with both the full ATA and human annotators. Across nearly all criteria, the ablated variant produces scores that are intermediate between annotator averages and the full ATA, but in a systematically misaligned way. For instance, in the travel-planning setting, ablation lowers the constraint-handling score from 3.53 to 3.17, pushing it further away from human judgment (4.07). Similarly, for Wikipedia writing, ablation sharply under-scores citation use (2.24) relative to both humans (3.53) and the full ATA (3.60). These gaps indicate that removing code analysis and literature search causes the system to lose sensitivity to criteria that hinge on factual grounding or structural reasoning.

\paragraph{Interpretation.}
Interestingly, the ablated ATA sometimes converges closer to human ratings in aggregate utility (e.g., 3.44 for travel vs.\ 3.36 for full ATA), but this reflects variance and calibration errors rather than more faithful evaluation. The full ATA consistently captures weaknesses that humans also penalize, whereas the ablation oscillates between being too lenient (e.g., Wikipedia context retention, 8.5 vs.\ 4.3) and overly harsh (e.g., travel constraint handling, 4.1 vs.\ 6.3) in earlier tables. These trends reinforce that the ablated variant drifts toward generic scoring, while the full ATA achieves stronger alignment with human assessments by anchoring its probes in evidence-driven test generation.

\section{Conclusions}
We introduce the \emph{Agent-Testing Agent (ATA)}, a meta-agent that automates adversarial evaluation of conversational agents. The ATA performs static code analysis of the agent under test, interactively elicits missing assumptions from the designer, mines related research and datasets, and then synthesizes persona-driven dialogues whose difficulty adaptively escalates based on judge feedback. Dialogues are scored with an LLM-as-a-Judge rubric and aggregated into quantitative metrics and qualitative bug reports suitable for regression testing and developer triage.

Empirically, testing a formal travel-planning agent and a Wikipedia-style article writer shows that the ATA surfaces overlapping functional weaknesses with human annotators while also discovering deeper, capability-level failures that humans do not consistently exercise. Humans emphasize tone and interpersonal quality more strongly, whereas ATA applies more mechanical, rubric-aligned pressure on end-to-end task success—yielding complementary coverage (cf. Tables~1–3). The ATA completes a full evaluation pass in roughly 20--30 minutes, whereas the ten-annotator round required days to schedule and execute, highlighting substantial time savings for iteration. 

An ablation without code analysis and web search reduces coverage and degrades calibration: scores become higher-variance and misaligned with both the full ATA and human judgments (e.g., severe under-scoring of citation/sourcing), underscoring the value of evidence-grounded weakness generation for high-impact evaluation. 

We release the full methodology and a production-ready, open-source implementation so that researchers and practitioners can reproduce our experiments, adapt the rubric to their domains, and continuously test their agents with minimal overhead. Looking forward, we see opportunities to expand to collaborative settings (multi-agent coordination), enrich pragmatic and stylistic criteria that humans detect well, and broaden domain adapters and tool simulators while preserving the ATA’s zero-domain-annotation setup.

\section*{Acknowledgments}

We thank Jenny Bellik for outlining and establishing the annotator guidelines, the first mockup of the annotation UI, and the rubric items. We thank Hannah Robertson and Rohit Jonnalagadda for coordinating and conducting the evaluation with the contracted annotators. Many thanks to John Blatz for fruitful discussions and his support.

\clearpage
\bibliographystyle{acl}
\bibliography{references}

\clearpage
\appendix
\section{Rubrics}
\label{rubrics}
In this appendix, we provide the detailed human-evaluation rubrics that were used to score both the travel-planning and Wikipedia-writing agents. These rubrics define the scale and descriptors for each criterion, including \emph{overall utility}, task-specific criteria such as \emph{constraint handling} and \emph{user communication} for the travel domain, and \emph{citations}, \emph{completeness}, and \emph{style/organization} for Wikipedia article generation. The goal of including these full rubrics is to make transparent the standards against which both human annotators and the automatic test agent (ATA) judgments were compared.

\setcounter{table}{0}
\renewcommand{\thetable}{A\arabic{table}}

\subsection{Overall Utility Rubric}
\begin{table*}[t]
\centering
\caption{Overall-utility scale used by human annotators.}
\label{tab:app-overall-utility}
\begin{minipage}{0.92\textwidth}
\centering
\begin{tabular}{l p{0.75\textwidth}}
\hline
\textbf{Level} & \textbf{Definition} \\
\hline
Negative & Conversation does not address the user’s requirements; includes serious errors or major unsupported assumptions. \\
Neutral  & Barely addresses requirements; misses major aspects; may include prominent hallucinations. \\
Small    & Reasonable starting point but needs improvement; minor hallucinations or missed opportunities. \\
Large    & Appropriately addresses requirements with only minor missed opportunities. \\
Exceptional & Fully addresses needs; contextually aware with accurate detail and appropriate format. \\
\hline
\end{tabular}
\end{minipage}
\end{table*}

\subsection{Travel—Constraint Handling}
\begin{table*}[t]
\centering
\caption{Travel-planning rubric: Constraint handling.}
\label{tab:app-travel-constraint}
\begin{minipage}{0.92\textwidth}
\centering
\begin{tabular}{l p{0.75\textwidth}}
\hline
\textbf{Level} & \textbf{Descriptor} \\
\hline
Bad (-1) & Ignores requirements or proposes unrelated/off-topic options. \\
Meh (0)  & Partially adapts but misses multiple significant aspects. \\
Ok (1)   & Mostly adapts but still misses some significant aspect. \\
Good (2) & Adapts with only minor preference misses. \\
Exceptional (3) & Perfectly adapts and/or clearly explains why a requirement cannot be met. \\
\hline
\end{tabular}
\end{minipage}
\end{table*}

\subsection{Travel—User Communication}
\begin{table*}[t]
\centering
\caption{Travel-planning rubric: User communication quality.}
\label{tab:app-travel-communication}
\begin{minipage}{0.92\textwidth}
\centering
\begin{tabular}{l p{0.75\textwidth}}
\hline
\textbf{Level} & \textbf{Descriptor} \\
\hline
Bad (-1) & Disorganized, confusing, or misleading; omits key context or next steps. \\
Meh (0)  & Understandable but incomplete; important clarifications or confirmations are missing. \\
Ok (1)   & Generally clear with occasional ambiguities; some justifications may be thin. \\
Good (2) & Clear, concise, and helpful; proactively surfaces trade-offs and next steps. \\
Exceptional (3) & Highly clear and collaborative; anticipates needs and communicates constraints transparently. \\
\hline
\end{tabular}
\end{minipage}
\end{table*}

\subsection{Wikipedia—Citations and Sourcing}
\begin{table*}[t]
\centering
\caption{Wikipedia-writing rubric: Citations and sourcing.}
\label{tab:app-wiki-citations}
\begin{minipage}{0.92\textwidth}
\centering
\begin{tabular}{l p{0.75\textwidth}}
\hline
\textbf{Level} & \textbf{Descriptor} \\
\hline
Bad (-1) & Missing citations for factual claims; unreliable or inappropriate sources. \\
Meh (0)  & Some citations provided but key claims lack support or sources are weak. \\
Ok (1)   & Most significant claims sourced; occasional gaps or borderline sources. \\
Good (2) & Consistently sources claims with appropriate, verifiable references. \\
Exceptional (3) & Fully compliant with sourcing guidelines; high-quality, diverse, and verifiable references. \\
\hline
\end{tabular}
\end{minipage}
\end{table*}

\subsection{Wikipedia—Completeness}
\begin{table*}[t]
\centering
\caption{Wikipedia-writing rubric: Completeness.}
\label{tab:app-wiki-completeness}
\begin{minipage}{0.92\textwidth}
\centering
\begin{tabular}{l p{0.75\textwidth}}
\hline
\textbf{Level} & \textbf{Descriptor} \\
\hline
Bad (-1) & Major sections missing; many core aspects of the topic are omitted. \\
Meh (0)  & Covers some sections but important aspects are thin or absent. \\
Ok (1)   & Adequate coverage of key sections with a few notable gaps. \\
Good (2) & Broad coverage with minor omissions; reasonable depth. \\
Exceptional (3) & Comprehensive coverage with appropriate granularity and balance. \\
\hline
\end{tabular}
\end{minipage}
\end{table*}

\subsection{Wikipedia—Style and Organization}
\begin{table*}[t]
\centering
\caption{Wikipedia-writing rubric: Style and organization.}
\label{tab:app-wiki-style}
\begin{minipage}{0.92\textwidth}
\centering
\begin{tabular}{l p{0.75\textwidth}}
\hline
\textbf{Level} & \textbf{Descriptor} \\
\hline
Bad (-1) & Disorganized; violates style conventions; inconsistent voice or formatting. \\
Meh (0)  & Basic structure present but sections are uneven or transitions are weak. \\
Ok (1)   & Mostly well-structured with occasional stylistic or organizational lapses. \\
Good (2) & Clear organization, consistent tone, and appropriate formatting. \\
Exceptional (3) & Highly readable, well-structured, and fully aligned with style guidelines. \\
\hline
\end{tabular}
\end{minipage}
\end{table*}

\section{Annotation UI}
\label{sec:annotation-ui}

\setcounter{figure}{0}
\renewcommand{\thefigure}{B\arabic{figure}}

\begin{figure*}[t]
\centering
\includegraphics[width=\textwidth]{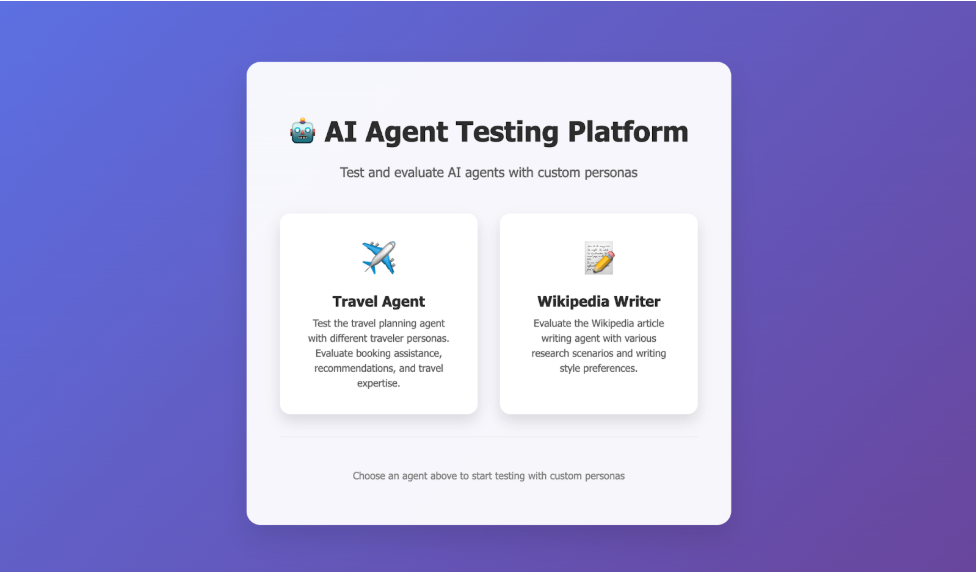}
\caption{\textbf{AI Agent Testing Platform landing page.} Entry point where annotators choose between the \emph{Travel Agent} and \emph{Wikipedia Writer} workflows before beginning persona setup and evaluation.}
\label{fig:annui-landing}
\end{figure*}

\begin{figure*}[t]
\centering
\includegraphics[width=\textwidth]{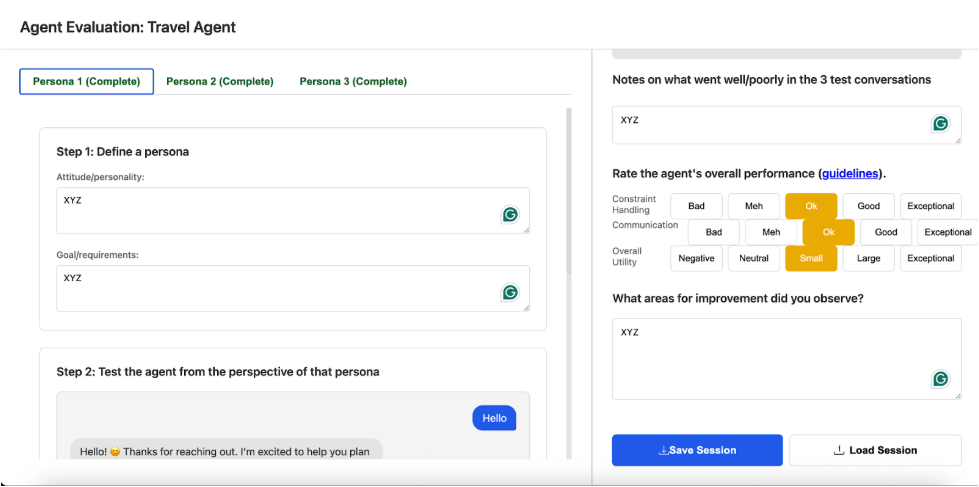}
\caption{\textbf{Travel Agent evaluation workspace.} Left: persona definition (attitude/goals) and chat window to test the agent from that persona’s perspective. Right: notes on successes/failures across the three test conversations, rubric grid for \emph{Constraint Handling}, \emph{Communication}, and \emph{Overall Utility}, plus an \emph{Areas for improvement} field.}
\label{fig:annui-travel}
\end{figure*}

\begin{figure*}[t]
\centering
\includegraphics[width=\textwidth]{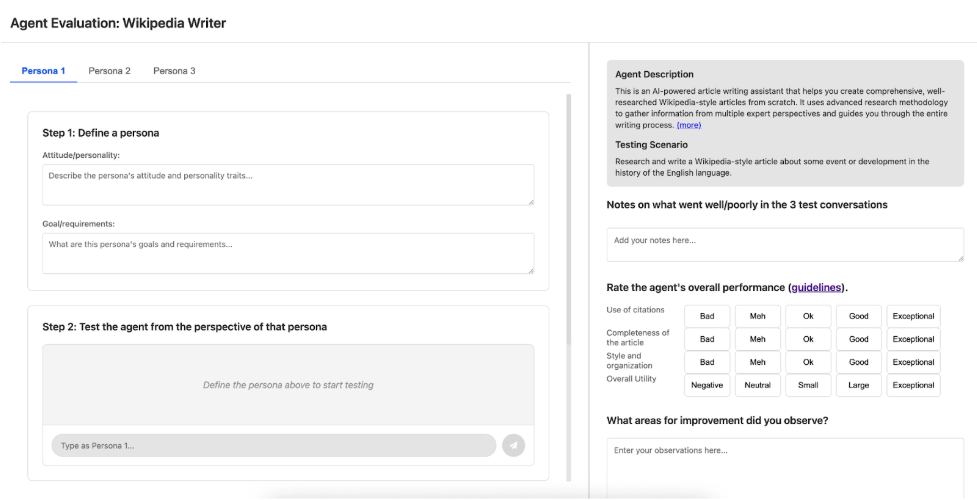}
\caption{\textbf{Wikipedia Writer evaluation workspace.} Left: persona setup and chat to test the agent. Right: task description and notes panel, with rubric grid for \emph{Citations}, \emph{Completeness}, \emph{Style/Organization}, and \emph{Overall Utility}, plus \emph{Areas for improvement}.}
\label{fig:annui-wiki}
\end{figure*}

This section documents the annotation interface used by human evaluators. 
The platform’s landing screen (Figure~\ref{fig:annui-landing}) lets annotators choose between the \emph{Travel Agent} and \emph{Wikipedia Writer}. 
Within each task-specific workspace, annotators (i) define three personas with attitudes and goals, (ii) conduct three short conversations from each persona’s perspective, and (iii) record structured notes and ratings. 
For the travel task (Figure~\ref{fig:annui-travel}), the right panel captures free-form notes on what went well/poorly across the three test conversations, a rubric-aligned grid for \emph{Constraint Handling}, \emph{Communication}, and \emph{Overall Utility}, and an \emph{Areas for improvement} field. 
For the Wikipedia task (Figure~\ref{fig:annui-wiki}), the layout mirrors travel but the rubric grid targets \emph{Citations}, \emph{Completeness}, \emph{Style/Organization}, and \emph{Overall Utility}. 
Both workspaces support saving and loading sessions to maintain consistency across annotators.

\end{document}